\documentclass[final]{cvpr}

\usepackage{times}
\usepackage{epsfig}
\usepackage{graphicx}
\usepackage{amsmath}
\usepackage{amssymb}
\usepackage{multirow}
\usepackage{subfigure}
\usepackage{placeins}
\usepackage{enumitem}
\usepackage{colortbl}
\usepackage[usenames,dvipsnames]{xcolor}
\usepackage{xcolor,colortbl}
\usepackage{xfrac}
\usepackage{dblfloatfix}
\usepackage{flushend}
\usepackage{verbatim}
\usepackage{booktabs}

\usepackage{xcolor}
\usepackage{textcomp}
\usepackage{color}
\usepackage{colortbl}
\usepackage{soul}

\definecolor{yellow}{rgb}{1,1,0.8}
\definecolor{green}{rgb}{0.7,0.95,0.65}
\definecolor{gray}{rgb}{0.9,0.9,0.9}
\definecolor{lightyellow}{rgb}{1,1, 0.8}

\definecolor{smurf}{rgb}{0.0863,0.5176,0.851} %

\def\rot#1{\rotatebox{90}{#1}}

\newcommand{\abs}[1]{\left\lvert#1\right\rvert}
\newcommand{\mean}[1]{\frac{1}{HW} \sum #1}

\newcommand{\lossfun}[1]{\mathcal{L}_\mathit{#1}}

\newcommand{\thispaperstitle}{\setulcolor{smurf} {\color{smurf} SMURF:} \ul{S}elf-Teaching \ul{M}ulti-Frame\\\ul{U}nsupervised \ul{R}AFT with \ul{F}ull-Image Warping \setulcolor{black}}

\newcommand{\tightparagraph}{\vspace{-0.35cm}\paragraph}

\usepackage[pagebackref=true,breaklinks=true,colorlinks,bookmarks=false,linkcolor=smurf,urlcolor=smurf]{hyperref}

\def\cvprPaperID{8823} %
\def\confYear{CVPR 2021}

\usepackage{etoolbox}
\newcommand{\zerodisplayskips}{%
  \setlength{\abovedisplayskip}{4pt}%
  \setlength{\belowdisplayskip}{6pt}%
  \setlength{\abovedisplayshortskip}{4pt}%
  \setlength{\belowdisplayshortskip}{6pt}}
\appto{\normalsize}{\zerodisplayskips}
\appto{\small}{\zerodisplayskips}
\appto{\footnotesize}{\zerodisplayskips}

\begin{document}

\title{\vspace{-0.5cm}\thispaperstitle\vspace{-0.25cm}} %

\def\aspace{\hspace{0.65cm}}

\newcommand\blfootnote[1]{%
  \begingroup
  \renewcommand\thefootnote{}\footnote{#1}%
  \addtocounter{footnote}{-1}%
  \endgroup
}

\author{Austin Stone$^{*1}$ \aspace Daniel Maurer$^{*2}$ \aspace Alper Ayvaci$^2$ \aspace Anelia Angelova$^1$ \aspace Rico Jonschkowski$^1$\\
$^1$Robotics at Google \hspace{4.5cm} $^2$Waymo\\
{\tt\small \{austinstone, anelia, rjon\}@google.com \aspace \{maurerd, ayvaci\}@waymo.com}\\
{\tt\small \url{https://github.com/google-research/google-research/tree/master/smurf}}
}

\maketitle
\thispagestyle{empty}

\blfootnote{$^{*}$Authors contributed equally.}

\begin{abstract}
We present SMURF, a method for unsupervised learning of optical flow that improves state of the art on all benchmarks by $36\%$ to $40\%$ (over the prior best method UFlow) and even outperforms several supervised approaches such as PWC-Net and FlowNet2. Our method integrates architecture improvements from supervised optical flow, i.e. the RAFT model, with new ideas for unsupervised learning that include a sequence-aware self-supervision loss, a technique for handling out-of-frame motion, and an approach for learning effectively from multi-frame video data while still only requiring two frames for inference.
\end{abstract}

\section{Introduction}
Optical flow describes a dense pixel-wise correspondence between two images, specifying for each pixel in the first image, where that pixel is in the second image. The resulting vector field of relative pixel locations represents apparent motion or ``flow'' between the two images. Estimating this flow field is a fundamental problem in computer vision and any advances in flow estimation benefit many downstream tasks such as visual odometry, multi-view depth estimation, and video object tracking.

Classical methods formulate optical flow estimation as an optimization problem~\cite{Brox04,Horn1981}. They infer, for a given image pair, a flow field that maximizes smoothness and the similarity of matched pixels. Recent supervised learning approaches instead train deep neural networks to estimate optical flow from examples of ground-truth annotated image pairs~\cite{Flownet2,Sun2018PWCNet,RAFT,yang2019volumetric}. Since obtaining ground truth flow is extremely difficult for real images, supervised learning is generally limited to synthetic data~\cite{FlowNet,FlyingThings}. While these methods have produced excellent results in the training domain, generalization is challenging when the gap between the target domain ant the synthetic training data is too wide.

Unsupervised learning is a promising direction to address this issue as it allows training optical flow models from unlabeled videos of any domain. The unsupervised approach works by combining ideas from classical methods and supervised-learning -- training the same neural networks as in supervised approaches but optimizing them with objectives such as smoothness and photometric similarity from classical methods. Unlike those classical methods, unsupervised approaches perform optimization not per image pair but jointly for the entire training set.

Since unsupervised optical flow takes inspiration from classical and supervised learning methods, we can make substantial progress by properly combining novel ideas with insights from these two directions. In this paper, we do exactly that and make the following three contributions:
\begin{enumerate}[leftmargin=*,itemindent=0pt,parsep=0pt,topsep=3pt]
\item We integrate the current best supervised model, RAFT~\cite{RAFT} with unsupervised learning and perform key changes to the loss functions and data augmentation to properly regularize this model for unsupervised learning.
\item We perform unsupervised learning on image crops while using the full image to compute unsupervised losses. This technique, which we refer to as full-image warping, improves flow quality near image boundaries.
\item We leverage a classical method for multi-frame flow refinement~\cite{maurer2018proflow} to generate better labels for self-supervision from multi-frame input. This technique improves performance especially in occluded regions without requiring more than two frames for inference.
\end{enumerate}

Our method \emph{\underline{S}elf-Teaching \underline{M}ulti-frame \underline{U}nsupervised \underline{R}AFT with \underline{F}ull-Image Warping} (SMURF) combines these three contributions and improves the state of the art (SOTA) in unsupervised optical flow in all major benchmarks, i.e. it reduces errors by 40 / 36 / 39 \% in the Sintel Clean / Sintel Final / KITTI 2015 benchmarks relative to the prior SOTA set by UFlow~\cite{jonschkowski2020matters}. These improvements also reduce the gap to supervised approaches, as SMURF is the first unsupervised optical flow method that outperforms supervised FlowNet2~\cite{Flownet2} and PWC-Net~\cite{Sun2018PWCNet} on all benchmarks.

\section{Related Work}

Optical flow was first studied in psychology to describe motion in the visual field~\cite{Gibson1950}. Later, the rise of computing in the 1980s led to analytical techniques to estimate optical flow from images~\cite{Horn1981,lucas1981iterative}. These techniques introduced photometric consistency and smoothness assumptions. These early methods do not perform any learning but instead solve a system of equations to find flow vectors that minimize the objective function for a given image pair. Follow-up work continued to improve flow accuracy, e.g. through better optimization techniques and occlusion reasoning~\cite{Brox04,Sun2010}.

Machine learning has helped to improve results substantially. First approaches used supervised convolutional neural networks that had relatively little flow-specific structure~\cite{FlowNet, Flownet2}. Others introduced additional inductive biases from classical approaches such as coarse-to-fine search~\cite{spynet2017,Sun2018PWCNet,yang2019volumetric}. The current best network architecture is the Recurrent All-Pairs Field Transforms (RAFT) model. It follows classical work that breaks with the coarse-to-fine assumption~\cite{chen2016full,steinbrucker2009large,xu2017accurate} and computes the cost volume between all pairs of pixels and uses that information to iteratively refine the flow field~\cite{RAFT}. All supervised methods rely heavily on synthetic labeled data for training. While producing excellent results in the supervised setting, RAFT has not previously been used in unsupervised learning. 

Unsupervised approaches appeared after supervised methods and showed that even without labels, deep learning can greatly outperform classical flow methods~\cite{im2020unsupervised,Janai2018ECCV,jonschkowski2020matters,liu2020learning,DDFlow,SelFlow,meister2018unflow,ranjan2019cvpr,ren2017unsupervised,wang2018unos,wang2018occlusion,yin2018geonet,jjyu2016unsupflow,Zhong2019UnsupervisedDE,zou2018dfnet}. Besides being more accurate than classical methods, learned methods are also faster at inference because all optimization occurs during training instead of during inference time.\footnote{The reason why learned unsupervised methods outperform classical methods is not obvious as both approaches optimize similar objectives. We speculate that the network structure introduces helpful inductive biases and that optimizing a single network for all training images provides a better regularization than independently optimizing a flow field for each image.}

A recent study~\cite{jonschkowski2020matters} performed an extensive comparison of the many proposed advances in unsupervised flow estimation and amalgamated these different works into a state of the art method called UFlow. We take this work as a starting point and build on the techniques suggested there such as range-map based occlusion reasoning~\cite{wang2018occlusion}, the Census loss for photometric consistency~\cite{meister2018unflow,zabih1994non}, edge aware smoothness~\cite{tomasi1998bilateral}, and self supervision~\cite{DDFlow,SelFlow}. 

We build on and substantially extend this prior work by enabling the RAFT model to work in an unsupervised learning setting through changes in the loss function and data augmentation. We also propose full-image warping to make the photometric loss useful for pixels that leave the (cropped) image plane. And we utilize a flow refinement technique~\cite{maurer2018proflow} to leverage multi-frame input during training to self-generate improved labels for self-supervision.

\section{SMURF for Unsupervised Optical Flow}

This section describes our method \emph{\underline{S}elf-Teaching \underline{M}ulti-frame \underline{U}nsupervised \underline{R}AFT with \underline{F}ull-Image Warping} (SMURF) in two parts. The first part covers preliminaries, such as the problem definition and components adopted from prior work. The second part describes the three main improvements in our method.

\subsection{Preliminaries on Unsupervised Optical Flow}

Given a pair of RGB images, $I_{1}, I_{2} \in \mathbb{R}^{H \times W \times 3}$, we want to estimate the flow field $V_{1}\in \mathbb{R}^{H \times W \times 2}$, which for each pixel in $I_{1}$ indicates the offset of its corresponding pixel in $I_{2}$. We address this in a learning-based approach where we want to learn a function $f_\theta$ with parameters $\theta$ that estimates the flow field for any image pair, such that $V_{1} = f_\theta(I_{1}, I_{2})$. We learn the parameters $\theta$ from data of unlabeled image sequences $D=\{(I_{i})_{i=1}^{m}\}$ by minimizing a loss function $\lossfun{}$, $\theta^*=\arg\min \lossfun{}(D, \theta)$.

We build on the loss function $\lossfun{}$ from UFlow~\cite{jonschkowski2020matters}, 
\begin{align*}
\lossfun{}(D, \theta) &=  \omega_{\mathit{photo}} \lossfun{photo}(D, \theta)\\&+ \omega_{\mathit{smooth}} \lossfun{smooth}(D, \theta) + \omega_{\mathit{self}} \lossfun{self}(D, \theta)\,,
\label{eq:total_loss}
\end{align*}
which is a weighted combination of three terms: occlusion aware photometric consistency $\lossfun{photo}$, edge-aware smoothness $\lossfun{smooth}$, and self-supervision $\lossfun{self}$.

The photometric consistency term is defined as
\begin{equation*}
    \lossfun{photo}(D, \theta) = \mean O_1 \odot \rho\Big(I_1, w\big(I_2, V_1\big)\Big)\,,
\end{equation*}
where $\odot$ represents element-wise multiplication and $\mean$ is shorthand notation for the mean over all pixels; we omit indices for better readability. The function $w(\cdot,\cdot)$ warps an image with a flow field, here estimated as $V_{1} = f_\theta(I_{1}, I_{2})$. The function $\rho(\cdot,\cdot)$ measures the photometric difference between two images based on a soft Hamming distance on the Census-transformed images and applies the generalized Charbonnier function \cite{meister2018unflow}. The occlusion mask $O_{1} \in \mathbb{R}^{H \times W}$ with entries $\in [0,1]$ deactivates photometric consistency for occluded locations. This is crucial because occluded pixels, by definition, are not visible in the other image and therefore need not obey appearance consistency -- the correct flow vector for an occluded pixel may point to a dissimilar-looking pixel in the other image. Like UFlow, we use a range-map based occlusion estimation~\cite{wang2018occlusion} with gradient stopping for all datasets except for KITTI, where forward backward consistency~\cite{Brox04} seems to yield better results~\cite{jonschkowski2020matters}. 

The $k$-th order edge-aware smoothness term is defined as
\resizebox{\linewidth}{!}{
\begin{minipage}{\linewidth}
\begin{align*}
\lossfun{smooth(k)}(D, \theta) &=
\mean{\Bigg( \exp {\left(- \frac{\lambda}{3} \sum_c \abs{\frac{\partial I_{1_c}}{\partial x}}\right)} \odot \abs{ \frac{\partial^k V_1}{\partial x^k} } \\
& + \exp {\left(-\frac{\lambda}{3}\sum_c\abs{\frac{\partial I_{1_c}}{\partial y}}\right)} \odot \abs{\frac{\partial^k V_1}{\partial y^k}}}\Bigg)\,. 
\end{align*}
\vspace{0cm}%
\end{minipage}}
The expression $\mean$ again computes the mean over all pixels, here for two components that compute the $k$-th derivative of the flow field $V_1$ along the $x$ and $y$ axes. These derivatives are each weighted by an exponential of the mean image derivative across color channels $c$. This weighting provides edge-awareness by ignoring non-smooth transitions in the flow field at visual edges~\cite{tomasi1998bilateral}. The sensitivity to visual edges is controlled by $\lambda$. 

The self-supervision term is defined as\\
\resizebox{\linewidth}{!}{
\begin{minipage}{\linewidth}
\begin{align*}
    \lossfun{self}(D, \theta) = \mean \hat{M}_1 \odot \Big(1-M_1\Big) \odot c\Big(\hat{V}_1, V_1\Big)\,.
\end{align*}
\vspace{0cm}%
\end{minipage}}
where a self-generated flow label $\hat{V}_1$~\cite{DDFlow} is compared to the flow prediction $V_1$ using the generalized Charbonnier function~\cite{Sun2010} $c(A, B) = \left((A - B)^2 + \epsilon^2 \right)^\alpha$ with all operations being applied elementwise. Our experiments use $\epsilon=0.001$ and $\alpha=0.5$. Prior work has modulated this comparison by masks $\hat{M}$ and $M$ for the generated labels and the predicted flow respectively, which are computed with a forward-backward
consistency check~\cite{Brox04}, so that self-supervision is only applied to image regions where the label is forward-backward consistent but the prediction is not. However, as we will describe below, our method actually improves when we remove this masking. 

The main reason to perform self-supervision is to improve flow estimation near the image boundaries where pixels can easily get occluded by moving outside of the image frame~\cite{DDFlow}. To address this problem via self-supervision, we apply the model on an image pair to produce a flow label and use this self-generated label to supervise flow predictions from a cropped version of the image pair. This way the model can transfer its own predictions from an easier to a more difficult setting. We build on and extend this technique with additional augmentations as described below.

\subsection{SMURF's Improvements}

After having covered the foundation that our method builds on, we will now explain our three major improvements: 1) enabling the RAFT architecture~\cite{RAFT} to work with unsupervised learning, 2) performing full-image warping while training on image crops, and 3) introducing a new method for multi-frame self-supervision.

\subsubsection{Unsupervised RAFT}
\label{sec:unsupervised_raft}

As model $f_\theta$ for optical flow estimation, we use the Recurrent All-Pairs Field Transforms (RAFT)~\cite{RAFT} model, a recurrent architecture that has achieved top performance when trained with supervision but has not previously been used with unsupervised learning. RAFT works by first generating convolutional features for the two input images and then compiling a 4D cost volume $C\in\mathbb{R}^{H\times W\times H \times W}$ that contains feature-similarities for all pixel pairs between both images. This cost volume is then repeatedly queried and fed into a recurrent network that iteratively builds and refines a flow field prediction. The only architectural modification we make to RAFT is to replace batch normalization with instance normalization~\cite{instancenorm} to enable training with very small batch sizes. Reducing the batch size was necessary to fit the model and the more involved unsupervised training steps into memory. But more importantly, we found that leveraging RAFT's potential for unsupervised learning \emph{requires key modifications to the unsupervised learning method}, which we will discuss next.

\tightparagraph{Sequence Losses} In supervised learning of optical flow, it is common to apply losses not only to the final output but also to intermediate flow predictions~\cite{RAFT, Sun2018PWCNet, yang2019volumetric}. In unsupervised learning this is not typically done -- presumably because intermediate outputs of other models are often at lower resolutions which might not work well with photometric and other unsupervised losses~\cite{jonschkowski2020matters,DDFlow}. However, RAFT produces intermediate predictions at full resolution and does not pass gradients between prediction steps. Thus, we apply the unsupervised losses at the entire sequence of RAFT's intermediate predictions, which we found to be essential to prevent divergence and to achieve good results. Similarly to supervised methods, we exponentially decay the weight of these losses at earlier iterations~\cite{RAFT}. $\lossfun{sequence}= \sum_{i=1}^{n} \gamma^{n-i}\lossfun{i}$, where $n$ is the number of flow iterations, $\gamma$ is the decay factor, and $\lossfun{i}$ is the loss at iteration $i$. Our experiments use $n=12$ and $\gamma=0.8$. 

\begin{figure}[t]
    \centering
    \includegraphics[width=\columnwidth]{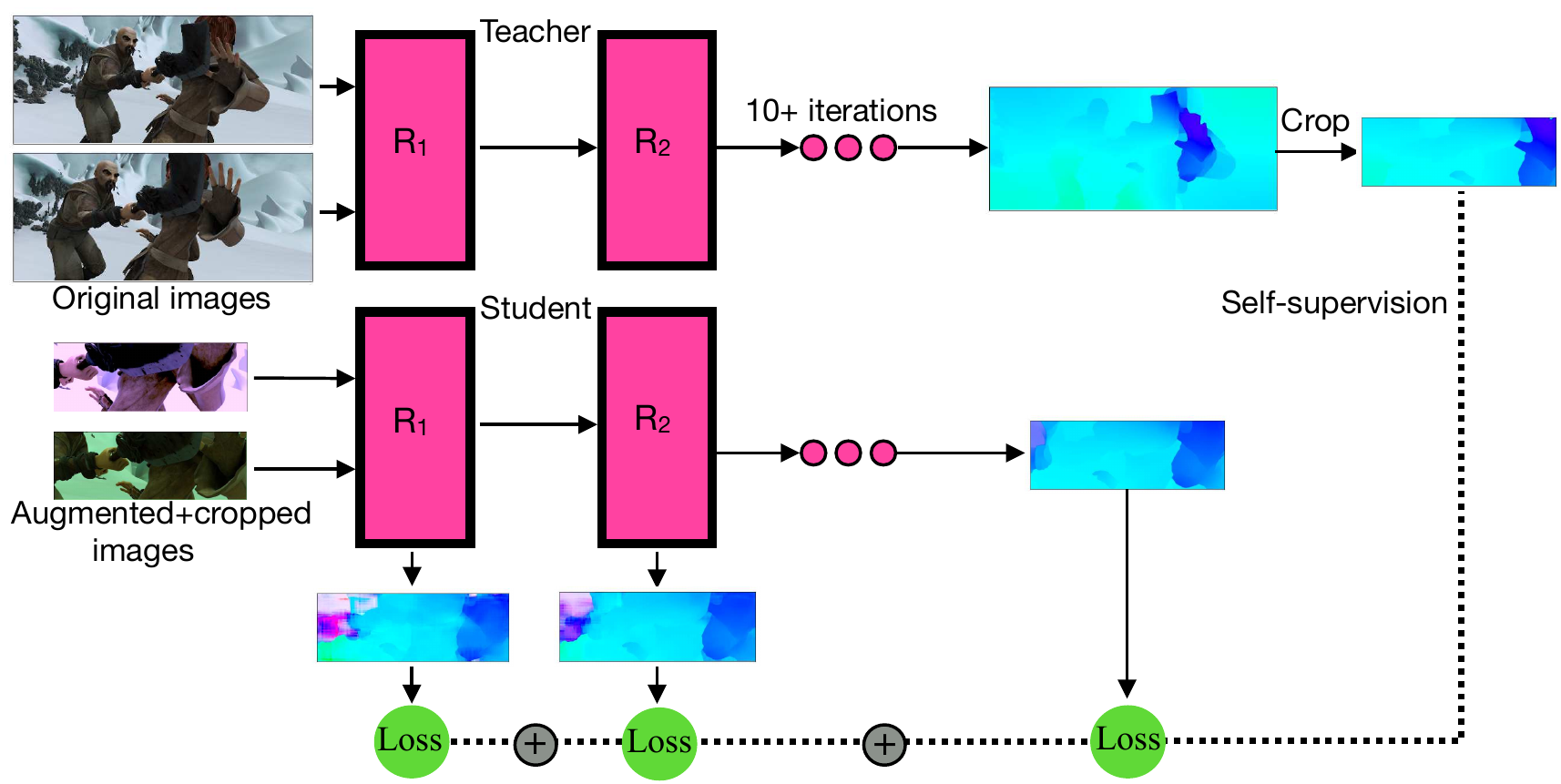}%
    \caption{Self-supervision with sequence loss and augmentation. We use a single model as both ``student'' and ``teacher''. As the teacher, we apply the model on full non-augmented images. As student, the model only sees a cropped and augmented version of the same images. The final output of the teacher is then cropped and used to supervise the predictions at all iterations of the student (to which the smoothness and photometric losses are applied as well). The advantages of this self-supervision method are threefold: (1) the model learns to ignore photometric augmentations (2) the model learns to make better predictions at the borders and in occluded areas of the image, and (3) early iterations of the recurrent model learn from the output at the final iteration.}
    \label{fig:model}
\end{figure}

\tightparagraph{Improved Self-Supervision} For self-supervision, we implement the sequence loss by applying the model to a full image, taking its final output, and using that to supervise all iterations of the model applied to the cropped image (see \autoref{fig:model}). Applying self-supervision from full to cropped images has been shown to provide a learning signal to pixels near the image border that move out of the image and therefore receive no signal from photometric losses~\cite{DDFlow}. Performing this self-supervision in a sequence-aware manner allows the earlier stages of the model to learn from the model's more refined final output. Potentially because of this inherent quality difference between the ``student'' and the ``teacher'' flow, we found that simplifying the self-supervision loss by removing masks $M$ and $\hat{M}$ improves performance further. The self-supervision loss without masks is $\lossfun{self}(D, \theta) = \mean c\Big(\hat{V}_1, V_1\Big)$. Note that this change does not affect the photometric loss $\lossfun{photo}$, where we continue to use $O$ to mask out occlusions.

\begin{figure}[t]
    \centering
    \includegraphics[width=\columnwidth]{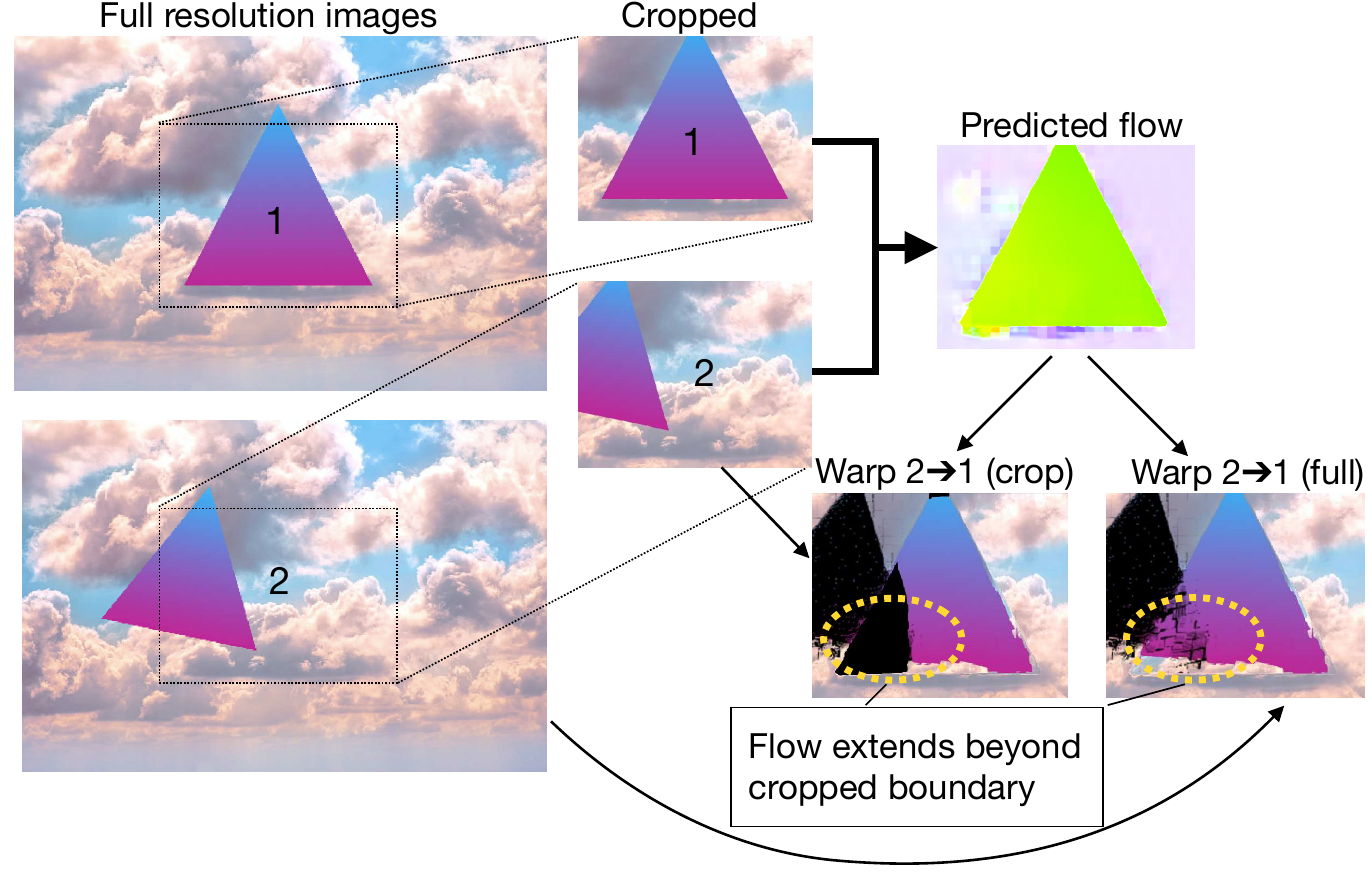}%
    \caption{Full-image warping. Images are cropped for flow prediction, but warping of image 2 with the predicted flow (to compute the photometric loss), is done with the full size image. The advantage is shown in the lower right: Compared to warping the cropped image (left), full-image warping reduces occlusions from out-of-frame motion (shown in black) and is able to better reconstruct image 1. When used during training, full-image warping provides a learning signal for pixels that move outside the cropped image boundary. The remaining occlusions in the reconstruction are due to noisy flow predictions for the out-of-frame motion.}
    \label{fig:full_warp}
\end{figure}

\tightparagraph{Extensive Data Augmentation} 
To regularize RAFT, we use the same augmentation as supervised RAFT~\cite{RAFT} which is much stronger than what has typically been used in unsupervised optical flow, except for the recent ARFlow~\cite{liu2020learning}. We randomly vary hue, brightness, saturation, stretching, scaling, random cropping, random flipping left/right and up/down, and we apply a random eraser augmentation that removes random parts of each image. All augmentations are applied to the model inputs, but not to the images used to compute the photometric and smoothness losses. The self-generated labels for self-supervision are computed from un-augmented images, which has the benefit of training the model to ignore these augmentations (see \autoref{fig:model}).

\begin{figure}[t]
    \centering
    \includegraphics[width=\columnwidth]{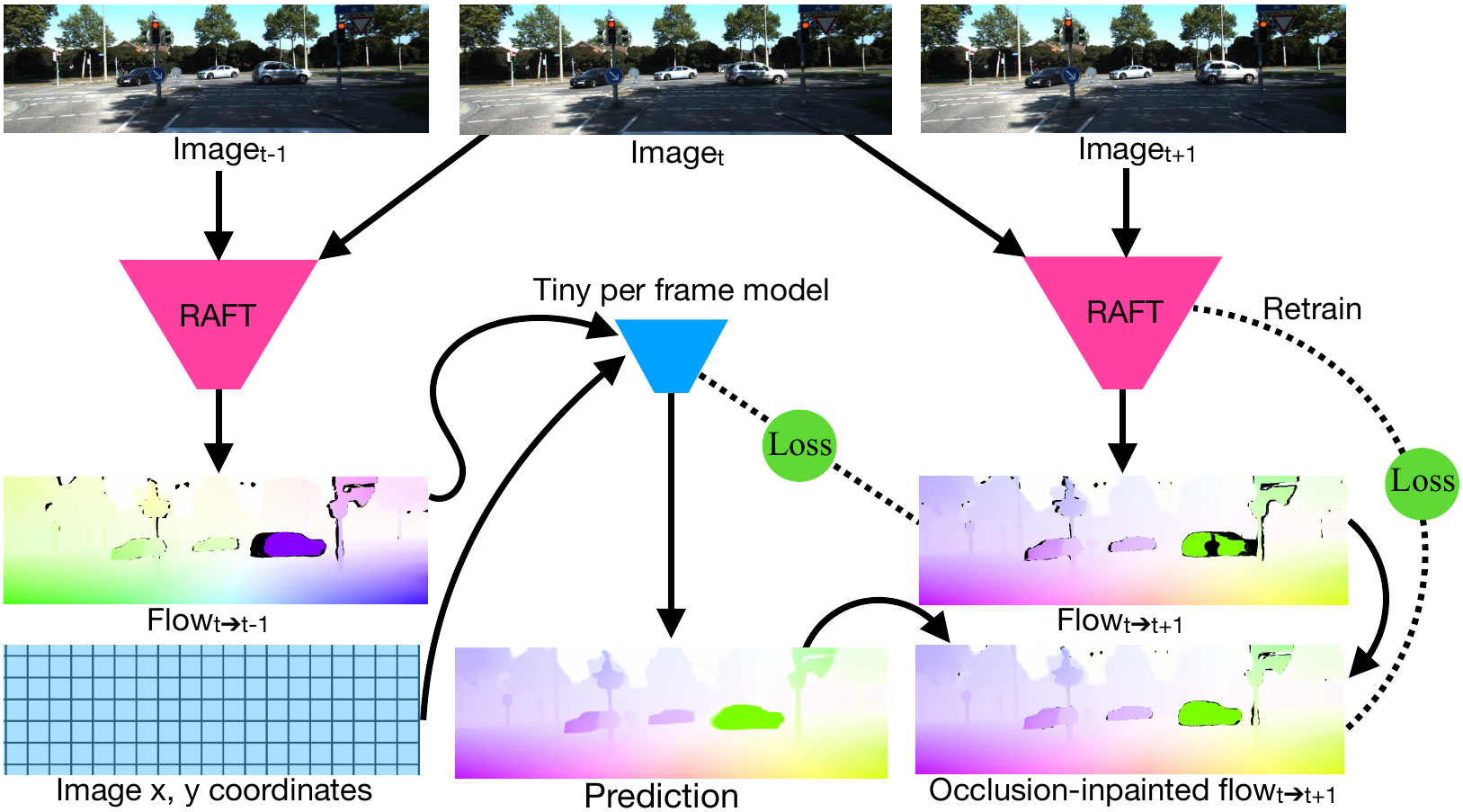}%
    \caption{Multi-frame self-supervision. From a sequence of three frames $t-1$, $t$, and $t+1$, we compute the backward flow ($t\rightarrow t-1$) and the forward flow ($t\rightarrow t+1$). The backward flow is then inverted via a tiny model (which is trained for this frame pair) and used to inpaint occluded regions in the forward flow. This inpainted flow field is then used to retrain the RAFT model.}
    \label{fig:multi_frame}
\end{figure}

\newcommand{\resultswidth}{1.06in}
\newcommand{\resultsspace}{\,}
\newcommand{\sidelabel}[1]{\rot{\footnotesize #1}}

\begin{figure*}
\centering
\resizebox{\textwidth}{!}{%
\begin{tabular}{@{}c@{\resultsspace}c@{\resultsspace}c@{\resultsspace}c@{\resultsspace}c@{\resultsspace}c@{}}
\includegraphics[width = \resultswidth]{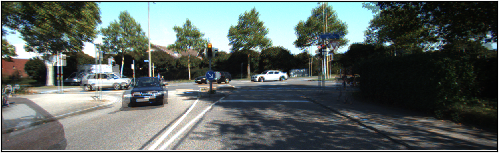} & 
\includegraphics[width = \resultswidth]{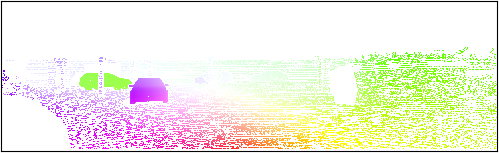} &
\includegraphics[width = \resultswidth]{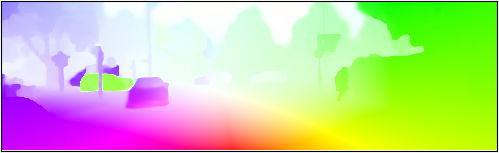} &
\includegraphics[width = \resultswidth]{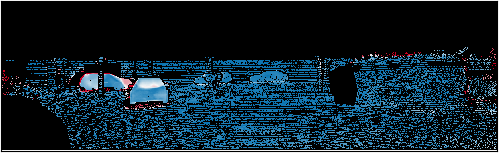} &
\includegraphics[width = \resultswidth]{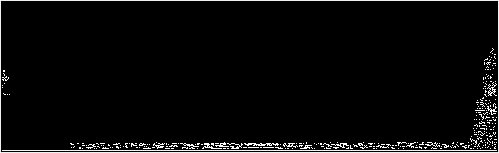} & 
\includegraphics[width = \resultswidth]{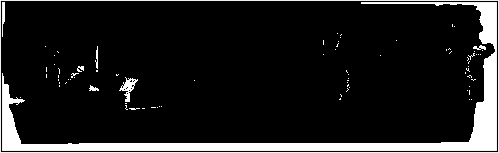} \\[-1mm]
\includegraphics[width = \resultswidth]{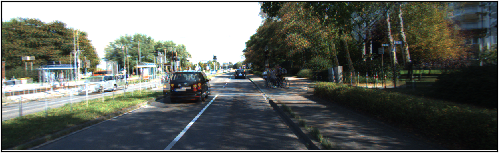} &
\includegraphics[width = \resultswidth]{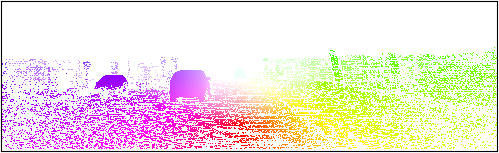} & 
\includegraphics[width = \resultswidth]{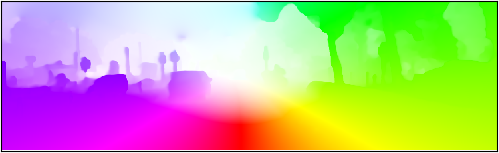} &
\includegraphics[width = \resultswidth]{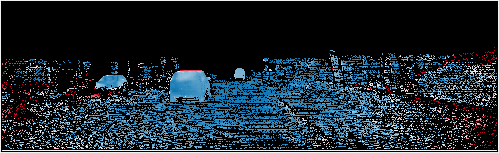} &
\includegraphics[width = \resultswidth]{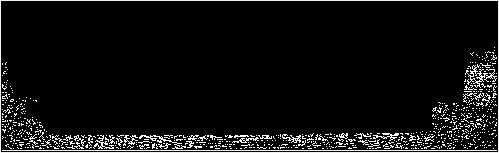} &
\includegraphics[width = \resultswidth]{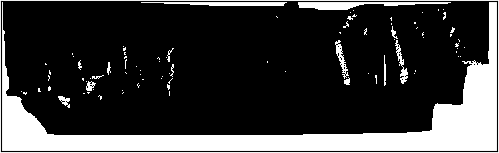} \\
\includegraphics[width = \resultswidth]{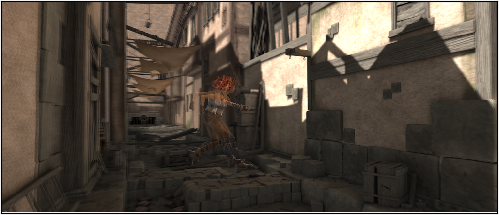} &
\includegraphics[width = \resultswidth]{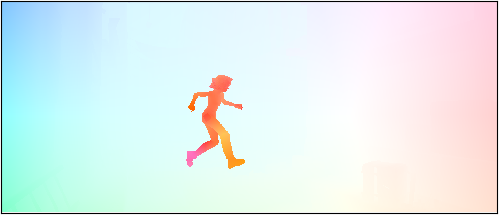} &
\includegraphics[width = \resultswidth]{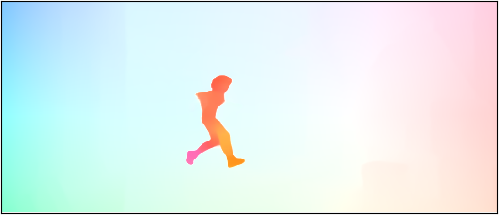} &
\includegraphics[width = \resultswidth]{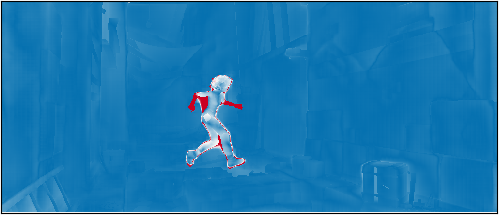} &
\includegraphics[width = \resultswidth]{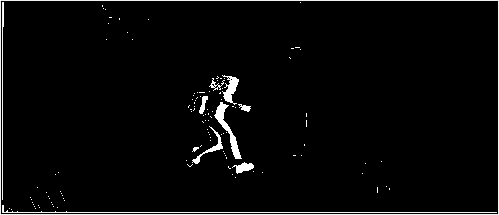} &
\includegraphics[width = \resultswidth]{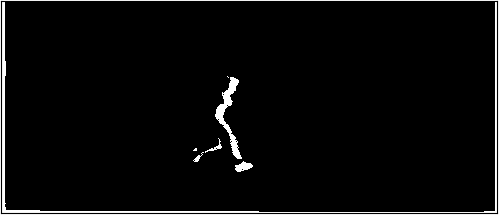} \\[-1mm]
\includegraphics[width = \resultswidth]{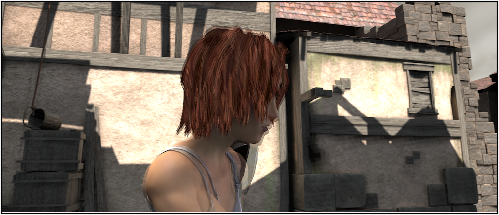} &
\includegraphics[width = \resultswidth]{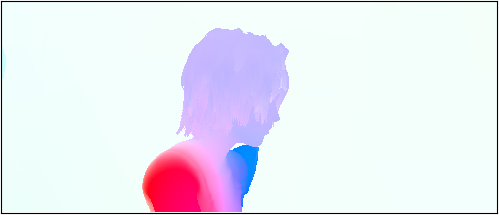} &
\includegraphics[width = \resultswidth]{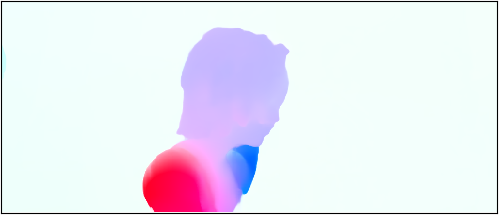} &
\includegraphics[width = \resultswidth]{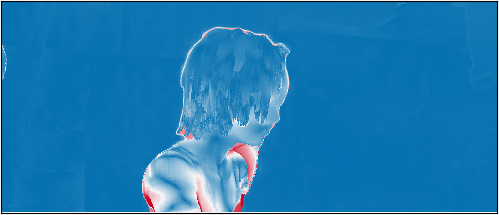} &
\includegraphics[width = \resultswidth]{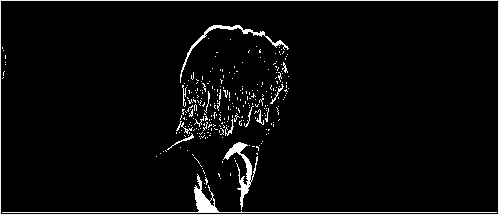} &
\includegraphics[width = \resultswidth]{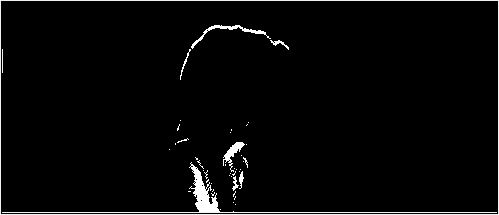} \\
\includegraphics[width = \resultswidth]{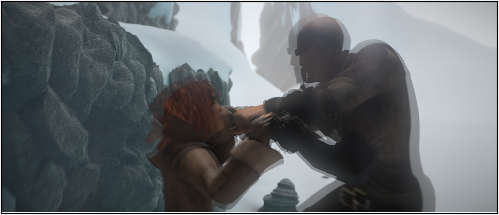} &
\includegraphics[width = \resultswidth]{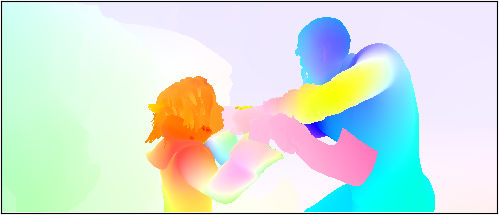} &
\includegraphics[width = \resultswidth]{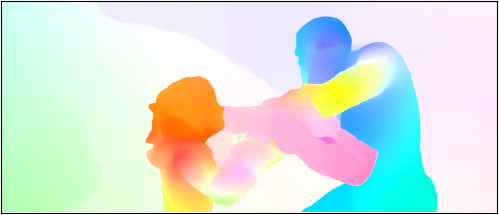} &
\includegraphics[width = \resultswidth]{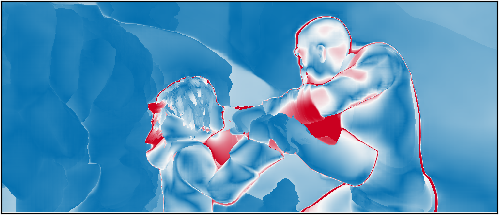} &
\includegraphics[width = \resultswidth]{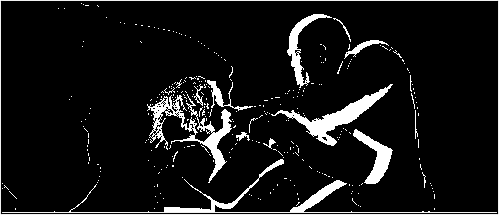} &
\includegraphics[width = \resultswidth]{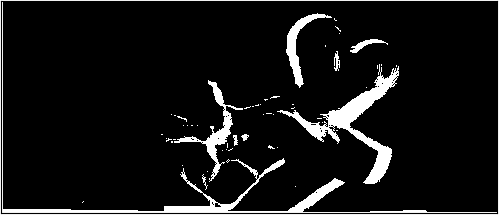} \\[-1mm]
\includegraphics[width = \resultswidth]{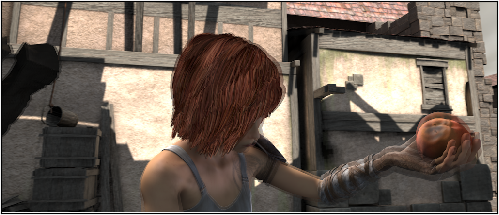} &
\includegraphics[width = \resultswidth]{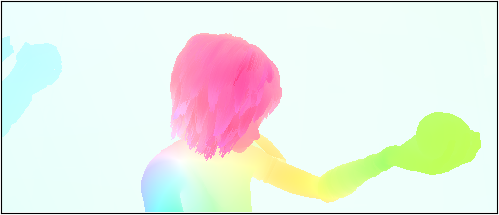} &
\includegraphics[width = \resultswidth]{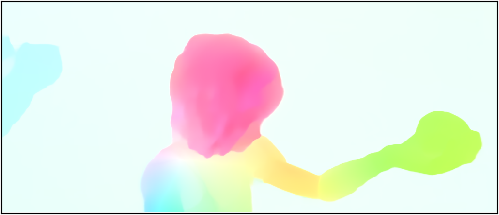} &
\includegraphics[width = \resultswidth]{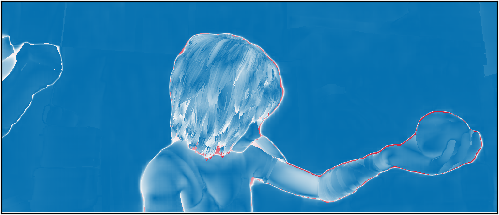} &
\includegraphics[width = \resultswidth]{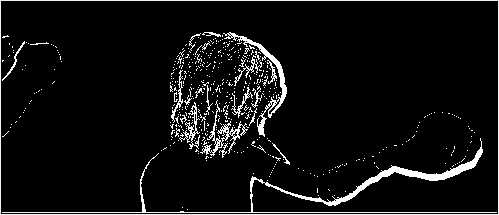} &
\includegraphics[width = \resultswidth]{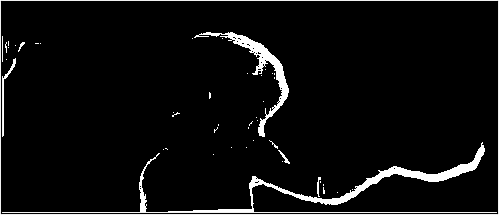} \\
{\footnotesize Overlayed Frames} & {\footnotesize True Flow} & {\footnotesize Predicted Flow} & {\footnotesize Flow Error} & {\footnotesize True Occlusions} & {\footnotesize Predicted Occlusions}\\[1mm]
\end{tabular} 

}
\caption{
Qualitative results for SMURF on two random examples from KITTI 2015, Sintel Clean, and Sintel Final (top to bottom) not seen during training. These results show the model's ability to estimate fast motions, relatively fine details, and substantial occlusions.
}
\label{fig:results}
\end{figure*}

\subsubsection{Full-Image Warping}
\label{sec:full_image_warping}
The photometric loss, which is essential for unsupervised optical flow estimation, is generally limited to flow vectors that stay inside the image frame because vectors that point outside of the frame have no pixels to compare their photometric appearance to. We address this limitation by computing the flow field from a cropped version of the images $I_1$ and $I_2$ while referencing the full, uncropped image $I_2$ when warping it with the estimated flow $V_1$ before computing the photometric loss (see \autoref{fig:full_warp}). As we also no longer mark these flow vectors that move outside the image frame as occluded, they now provide the model with a learning signal. We use full-image warping for all datasets except Flying Chairs, where we found that cropping the already small images hurt performance.

\subsubsection{Multi-Frame Self-Supervision}
\label{sec:multi_frame_self_supervision}

Finally, we propose to leverage multi-frame information for self-supervision to generate better labels in occluded areas, inspired by work that used a similar technique for inference~\cite{maurer2018proflow}. For multi-frame self-supervision, we take a frame $t$ and compute the forward flow to the next frame ($t\rightarrow t+1$) and the backward flow to the previous frame ($t\rightarrow t-1$). We then use the backward flow to predict the forward flow through a tiny learned inversion model and use that prediction to inpaint areas that were occluded in the original forward flow but were not occluded in the backward flow -- which is why the estimate from the backward flow is more accurate (see \autoref{fig:multi_frame}). The tiny model for the backward-forward inversion consists of three layers of 3$\times$3 convolutions with [16, 16, 2] channels that are applied on the backward flow and the image coordinates normalized to $[-1, 1]$. The model is re-initialized and trained per frame using the non-occluded forward flow as supervision, after which the in-painted flow field is stored and used for self-supervision. We apply multi-frame self-supervision only at the final stage of training. Importantly, we use multiple frames only during training and not for inference.

\section{Experiments}

\label{sec:experiments}

Our hyperparameters are based on UFlow~\cite{jonschkowski2020matters} with slight modifications based on further hyperparameter search. In all experiments, we use weights $\omega_{\mathit{photo}}=1$ and $\omega_{\mathit{self}}=0.3$. Regarding smoothness parameters, we set edge sensitivity $\lambda=150$ and $\omega_{\mathit{smooth}}=4$ for KITTI and Chairs and $\omega_{\mathit{smooth}}=2.5$ for Sintel. We use 2nd order smoothness ($k=2$) for KITTI and 1st order smoothness ($k=1$) for all other datasets. We train for 75K iterations with batch size 8, except for Sintel where we train for only 15K iterations to avoid overfitting. The self-supervision weight is set to 0 for the first 40\% of gradient steps, then linearly increased to $0.3$ during the next 10\% of steps and then kept constant. All of our training uses Adam~\cite{KingmaAdam} ($\beta_1=0.9$, $\beta_2=0.999$, $\epsilon=10^{-8}$) with learning rate of $0.0002$, which takes about 1 day to converge on 8 GPUs running synchronous SGD. On all datasets, the learning rate is exponentially decayed to $\frac{1}{1000}$ of its original value over the last 20\% of steps. On Sintel and on Flying Chairs we train with random crops of $368 \times 496$, and on KITTI we train with random crops of size $296 \times 696$. For all datasets we evaluated our model on the input resolution that had the lowest average end point error on the training set; $488 \times 1144$ for KITTI and $480 \times 928$ for Sintel. All test images were bilinearly resized to these resolutions during inference, and the resulting flow field was bilinearly resized and rescaled back to the native image size to compute evaluation metrics.

During the second stage of training when multi-frame self-supervision is applied, we generate labels for all images using the model trained according to the procedure described above. We then continue training the same model with only the self-supervision loss for an additional 30K iterations using the multiframe generated labels. We use the same hyperparameters as in the first stage but exponentially decay the learning rate for the last 5K iterations. For our best performing Sintel model, we train with the KITTI self-supervision labels mixed in at a ratio of 50\%.

\tightparagraph{Datasets}

\begin{table*}[t]
    \centering
    \resizebox{0.85\textwidth}{!}{%
	\begin{tabular}{ll cc c cc c cccc}
		 & & \multicolumn{2}{c}{Sintel Clean~\cite{ButlerECCV2012}} & \phantom{aaa} & \multicolumn{2}{c}{Sintel Final~\cite{ButlerECCV2012}} & \phantom{aaa} & \multicolumn{4}{c}{KITTI 2015~\cite{KITTI2015}} \\
		 \cmidrule{3-4} \cmidrule{6-7} \cmidrule{9-12}     
        & & \multicolumn{2}{c}{EPE} && \multicolumn{2}{c}{EPE} && EPE & EPE (noc) & \multicolumn{2}{c}{ER in \%}\\
        &Method& \textit{train} & \textit{test} && \textit{train} & \textit{test} && \textit{train} & \textit{train} & \textit{train} & \textit{test}\\
        \midrule
        \parbox[t]{2mm}{\multirow{5}{*}{\rotatebox[origin=c]{90}{\small{Supervised}}}}
        \parbox[t]{2mm}{\multirow{5}{*}{\rotatebox[origin=c]{90}{\footnotesize{in domain}}}}
        & FlowNet2-ft \cite{Flownet2} & (1.45) & 4.16 && (2.01) & 5.74 && (2.30) & -- & (8.61) & 11.48 \\
        & PWC-Net-ft \cite{Sun2018PWCNet} & (1.70) & 3.86 && (2.21) & 5.13 && (2.16) & -- & (9.80) &9.60 \\
        & SelFlow-ft \cite{SelFlow}~$^{\text{(MF)}}$ & (1.68) & [3.74] && (1.77) & \{4.26\} && (1.18) & -- & -- & 8.42 \\ 
        & VCN-ft \cite{yang2019volumetric}  & (1.66) & 2.81 && (2.24) & 4.40 && (1.16) & -- &  (4.10) & 6.30 \\
        & RAFT-ft \cite{RAFT}  & (0.76) & {\bf 1.94} && (1.22) & {\bf 3.18} && (0.63) & -- &  (1.5) & {\bf 5.10} \\
        \midrule
        \parbox[t]{2mm}{\multirow{4}{*}{\rotatebox[origin=c]{90}{\small{Supervised}}}}
        \parbox[t]{2mm}{\multirow{4}{*}{\rotatebox[origin=c]{90}{\footnotesize{out of domain}}}}
        & FlowNet2 \cite{Flownet2} & 2.02 & {\bf 3.96} && 3.14 & {\bf 6.02} && 9.84 & -- & 28.20 & -- \\
        & PWC-Net \cite{Sun2018PWCNet} & 2.55 & -- && 3.93 & -- && 10.35 & -- & 33.67 & -- \\
        & VCN \cite{yang2019volumetric} & 2.21 & -- && 3.62 & -- && 8.36 & -- & 25.10 & -- \\
        & RAFT \cite{RAFT} & {\bf 1.43} & -- && {\bf 2.71} & -- && {\bf 5.04} & -- & {\bf 17.4} & -- \\
        \midrule
        \parbox[t]{2mm}{\multirow{10}{*}{\rotatebox[origin=c]{90}{Unsupervised}}}
        & EPIFlow \cite{Zhong2019UnsupervisedDE} & 3.94 & 7.00 && 5.08 & 8.51 && 5.56 & 2.56 & -- & 16.95 \\ %
        & DDFlow \cite{DDFlow} & \{2.92\} & 6.18 && \{3.98\} & 7.40 && [5.72] & [2.73] & -- & 14.29 \\ %
        & SelFlow \cite{SelFlow}~$^{\text{(MF)}}$ & [2.88] & [6.56] && \{3.87\} & \{6.57\} && [4.84] & [2.40] & -- & 14.19  \\ %
        & UnsupSimFlow \cite{im2020unsupervised} & \{2.86\} & 5.92 && \{3.57\} & 6.92 && [5.19] & -- & -- & [13.38]\\
        & ARFlow \cite{liu2020learning}~$^{\text{(MF)}}$ & \{2.73\} & \{4.49\} && \{3.69\} & \{5.67\} && [2.85] & -- & -- & [11.79] \\
        & UFlow \cite{jonschkowski2020matters} & 3.01 & {5.21} && 4.09 & {6.50} && 2.84 & 1.96 & 9.39 & { 11.13} \\
        & SMURF-test (ours) & {\bf 1.99} &  -- && {\bf 2.80} & --  && {\bf 2.01} & {\bf 1.42} & {\bf 6.72} & -- \\
        & SMURF-train (ours) & \{1.71\} & {\bf 3.15} && \{2.58\} & {\bf 4.18} && \{2.00\} & \{1.41\} & \{6.42\} & {\bf 6.83} 
	\end{tabular}}
	\vspace{2.0mm}
    \caption{Comparison to state of the art. SMURF-train / test is our model trained on the train / test split of the corresponding dataset. The best results per category are shown in bold -- note that supervision ``in domain'' is often not possible in practice as flow labels for real images are difficult to obtain. Braces indicate results that might have overfit because evaluation data was used for training: ``()'' evaluated on the same labeled data as was used for training, ``\{\}'' trained on the unlabeled evaluation set, and ``[]'' trained on data highly related to the evaluation set (e.g., the entire Sintel Movie or $<5$ frames away from an evaluation image in KITTI). Methods that use multiple frames at inference are denoted with ``MF''; our method uses multiple frames only during training.}
    \label{tab:main}
\end{table*}

We train and evaluate our model according to the conventions in the literature using the following optical flow datasets: Flying Chairs~\cite{FlowNet}, Sintel~\cite{ButlerECCV2012} and KITTI 2015~\cite{KITTI2015}. We pretrain on Flying Chairs before fine tuning on Sintel or KITTI. Similar to UFlow~\cite{jonschkowski2020matters}, we did not find a benefit to pretraining on more out-of-domain data, e.g. Flying Things~\cite{FlyingThings}. None of the training techniques in our method uses any ground truth labels. We train on the ``training'' portion of Flying Chairs, and divide the Sintel dataset according to its standard train / test split. For KITTI, we train on the multi-view extension following the split used in prior work~\cite{jonschkowski2020matters,Zhong2019UnsupervisedDE}: We train two models, one on the multi-view extension of the training set and one on the extension of the test set, evaluate these models appropriately. For ablations, we report metrics after training on the ``test" portion of the dataset (which does not include labels) and evaluating on the training set, and for final benchmark numbers we report results after training on the training portion only. For our benchmark result on Sintel, we train on a 50\% mixture of the KITTI and Sintel multi-frame self-supervision labels. For all datasets we report endpoint error (``EPE''), and for KITTI, we additionally report error rates (``ER''), where a prediction is considered erroneous if its EPE is $>3$ pixels or $>5\%$ of the length of the true flow vector. We generally compute these metrics for all pixels, except for ``EPE (noc)'' where only non-occluded pixels are considered.

\section{Results}

In this section, we compare SMURF to related methods, ablate the proposed improvements, and show limitations.

\subsection{Comparison to State of the Art}

\begin{figure*}
    \centering 
    \resizebox{\linewidth}{!}{
    \includegraphics{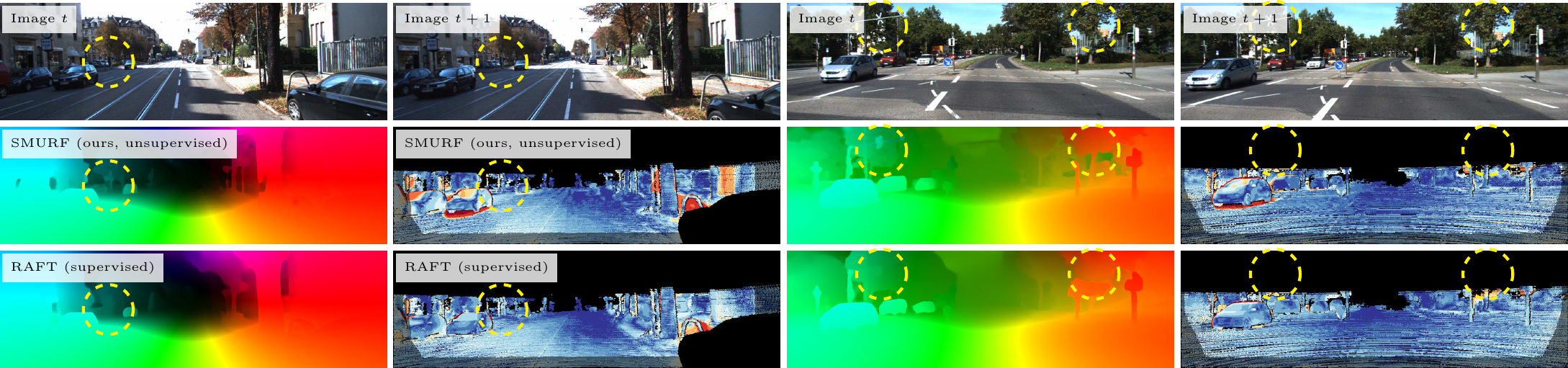}
    }
    \caption{Qualitative comparison of unsupervised SMURF and supervised RAFT. Maybe unsurprisingly supervised training in domain reduces errors especially for challenging and ambiguous cases, e.g. reflections and shadows of moving objects. But interestingly, there are also areas (highlighted) where our unsupervised method works better than the supervised alternative, apparently at small objects in places where labels are sparse and potentially imperfect (left) or non-existent (right).}
    \label{fig:kitti_supervised_vs_unsupervised}
\end{figure*}

Qualitative results for SMURF are shown in \autoref{fig:results} and a comparison to other methods can be found in \autoref{tab:main}. This comparison shows that our model substantially outperforms all prior published methods on unsupervised optical flow on all benchmarks. Compared to the the previous state of the art method UFlow~\cite{jonschkowski2020matters}, our method reduces benchmark test errors by 40 / 36 / 39 \% for Sintel Clean / Sintel Final / KITTI 2015. When we compare SMURF trained in domain to supervised methods trained out of domain, SMURF outperforms all supervised methods except RAFT~\cite{RAFT} on all benchmarks\footnote{This is a fair comparison as obtaining labels for a given domain is extremely difficult while our method trains on readily available video data.}. Compared to supervised RAFT, it performs a bit worse on Sintel (1.99 vs 1.43 on Sintel Clean, 2.80 vs. 2.71 on Sintel Final) but much better on KITTI 2015 (2.01 vs. 5.04 EPE, 6.72\% vs. 17.4\% ER). SMURF even outperforms some supervised methods when they are finetuned on the test domain, e.g. FlowNet2-ft~\cite{Flownet2}, PWC-Net-ft~\cite{Sun2018PWCNet}, and SelFlow-ft~\cite{SelFlow}. Only VCN-ft~\cite{yang2019volumetric} and RAFT-ft~\cite{RAFT} achieve better performance here. The inference time of our model is the same as the supervised RAFT model, approximately 500ms when using 12 recurrent iterations. A qualitative comparison to supervised RAFT is shown in \autoref{fig:kitti_supervised_vs_unsupervised}.

\begin{table}
\resizebox{\linewidth}{!}{
\centering
\begin{tabular}{llccccccccc}
&&\phantom{a}& Chairs & \phantom{a} & \multicolumn{2}{c}{Sintel \textit{train}} & \phantom{a} & \multicolumn{2}{c}{KITTI-15 \textit{train}} \\
\cmidrule{4-4} \cmidrule{6-7} \cmidrule{9-10} 
& Method && \textit{test} && Clean & Final && EPE & ER\%\\
\midrule
\parbox[t]{2mm}{\multirow{3}{*}{\rotatebox[origin=c]{90}{Train on}}}
\parbox[t]{2mm}{\multirow{3}{*}{\rotatebox[origin=c]{90}{Chairs}}}
& DDFlow~\cite{DDFlow} &&\cellcolor{gray} 2.97 && 4.83 & 4.85 && 17.26 &  -- \\
& UFlow~\cite{jonschkowski2020matters} && \cellcolor{gray} 2.55 && 4.36 & 5.12 && 15.68 & 32.69 \\
&  SMURF (ours) && \cellcolor{gray} 1.72 && {\bf 2.19} & {\bf 3.35} && {\bf 7.94} & {\bf 26.51} \\
\midrule
\parbox[t]{2mm}{\multirow{3}{*}{\rotatebox[origin=c]{90}{Train on}}}
\parbox[t]{2mm}{\multirow{3}{*}{\rotatebox[origin=c]{90}{Sintel}}} 
& DDFlow~\cite{DDFlow} && 3.46 && \cellcolor{gray} \{2.92\} & \cellcolor{gray} \{3.98\} && 12.69 & -- \\
&  UFlow~\cite{jonschkowski2020matters} && 3.25 && \cellcolor{gray} 3.01 & \cellcolor{gray} 4.09 && 7.67 & 17.41 \\
&  SMURF (ours) && {\bf 1.99} && \cellcolor{gray} 1.99 & \cellcolor{gray} 2.80 && {\bf 4.47} & {\bf 12.55} \\
\midrule
\parbox[t]{2mm}{\multirow{3}{*}{\rotatebox[origin=c]{90}{Train on}}} 
\parbox[t]{2mm}{\multirow{3}{*}{\rotatebox[origin=c]{90}{KITTI}}} 
& DDFlow~\cite{DDFlow} && 6.35 && 6.20 & 7.08 && \cellcolor{gray} [5.72] & \cellcolor{gray} -- \\
&  UFlow~\cite{jonschkowski2020matters} && 5.05 && 6.34 & 7.01 && \cellcolor{gray} 2.84 & \cellcolor{gray} 9.39 \\
&  SMURF (ours) && {\bf 3.26} && {\bf 3.38} & {\bf 4.47} && \cellcolor{gray} 2.01 & \cellcolor{gray} 6.72 \\
\end{tabular}}
\vspace{0.0mm}
\caption{Generalization across datasets. These results compare our method to state of the art unsupervised methods in a setting where a model is trained on one dataset and tested on different one.}
\label{table:generalization}
\end{table}

\begin{table}[t]
    \centering
    \resizebox{\linewidth}{!}{%
	\begin{tabular}{lcccccccc}
		 &\multicolumn{2}{c}{Trained on} && \multicolumn{2}{c}{KITTI-15 \emph{train}} && \multicolumn{2}{c}{KITTI-15 \emph{test}} \\
		 \cmidrule{2-3} \cmidrule{5-6} \cmidrule{8-9}
		 Method & Flow & Stereo & \phantom{a} & EPE & D1(all)\% & \phantom{a} & D1(noc)\% & D1(all)\% \\
        \midrule
        SGM \cite{Hirschmueller2008} &--&--&& -- & -- && 5.62 & 6.38 \\
        SsSMNet \cite{SsSMnet2017} &--&\checkmark&& -- & -- && \{3.06\} & \{3.40\} \\
        UnOS \cite{UnOS} &\checkmark&\checkmark&& -- & 5.94 && -- & 6.67 \\
        Flow2Stereo \cite{flow2stereo} &\checkmark&\checkmark&& 1.34 & 6.13 && 6.29 & 6.61 \\
        Reversing-PSMNet \cite{ReversingTheCycle} & -- & \checkmark && \bf 1.01 & \bf 3.85 && \bf 3.86 & \bf 4.06 \\
        SMURF-train (ours) &\checkmark&--&& 1.03 & 4.31 && 4.51 & 4.77
	\end{tabular}}
	\vspace{0.0mm}
    \caption{Stereo depth estimation. Without fine-tuning, our flow model estimates stereo depth ``zero-shot'' at an accuracy comparable to state of the art unsupervised methods trained for that task.}
    \label{tab:depth}
\end{table}

In \autoref{table:generalization}, we compare generalization across domains to prior unsupervised methods and find substantial improvements for every combination of training and test domain. Even after only training on Flying Chairs, our model already achieves an EPE of 2.19 on Sinel Clean and 3.35 on Sinel Final, which is superior to all prior unsupervised techniques even if fine tuned on Sintel data (first and third column in \autoref{tab:main}).

We also tested generalization from optical flow to stereo depth estimation. Here we evaluated our exact trained KITTI-15 flow model in the KITTI 2015 stereo depth benchmark~\cite{KITTI2015}. Despite neither tailoring the architecture or losses to stereo depth estimation nor training or fine tuning on any stereo data, our method achieves results that are competitive with the best unsupervised stereo depth estimation methods (see \autoref{tab:depth}).

\begin{table}
\centering
\resizebox{\linewidth}{!}{
    \begin{tabular}{lllcccccc}
    &&&\phantom{a}& \multicolumn{2}{c}{Sintel \textit{train}} & \phantom{a} & \multicolumn{2}{c}{KITTI-15 \textit{train}} \\
    \cmidrule{5-6} \cmidrule{8-9} 
    Method & Model & Training input && Clean & Final && EPE & ER\%\\
    \midrule
    UFlow~\cite{jonschkowski2020matters} & PWC & High res. && 3.01 & 4.09 && 2.84 & 9.39 \\
    UFlow & PWC & Low res. && 3.43 & 4.33 && 3.87 & 13.04 \\
    UFlow & RAFT & Low res. && 3.36 & 4.32 && 4.27$^{+}$ & 14.18$^{+}$ \\
    SMURF & PWC & Image crop && 2.63 & 3.66 && 2.73 & 9.33 \\
    SMURF & RAFT & Image crop && {\bf 1.99} & {\bf 2.80} && {\bf 2.01} & {\bf 6.72} \\
    \end{tabular}}
\vspace{0.0mm}
\caption{Test of whether replacing PWC-Net with RAFT improves the prior best unsupervised method UFlow. Due to memory constraints, training RAFT in this setting requires a lower resolution than UFlow~\cite{jonschkowski2020matters} (384$\times$512 for Sintel, 320$\times$704 for KITTI). We use that resolution for a side-by-side comparison to PWC (rows 2-3), which shows that RAFT performs poorly without our proposed modifications. Results marked with $^{+}$ overfit the dataset and generate worse performance than reported here at the end of training.}
\label{table:comparison_pwc_raft}
\end{table}

\begin{table}
\centering
\resizebox{0.9\linewidth}{!}{
    \begin{tabular}{cccccccccc}
    &&& &\phantom{a}& \multicolumn{2}{c}{Sintel \textit{train}} & \phantom{a} & \multicolumn{2}{c}{KITTI-15 \textit{train}} \\
    \cmidrule{6-7} \cmidrule{9-10} 
    SQ & AU & FW & MF && Clean & Final && EPE & ER\%\\
    \midrule
    -- & -- & -- & -- && 2.92 & 3.92 && 8.40 & 18.57\\
    -- & \checkmark & -- & -- && \multicolumn{2}{c}{\textit{--- diverged ---}} && \multicolumn{2}{c}{\textit{--- diverged ---}}\\
    \checkmark & -- & -- & -- && 2.66 & 3.86 && 5.01$^+$ & 17.50$^+$\\
    \checkmark & \checkmark & -- & -- && 2.38 & 3.17 && 3.64$^+$ & 13.05$^+$\\
    -- & \checkmark & \checkmark & -- && \multicolumn{2}{c}{\textit{--- diverged ---}} && \multicolumn{2}{c}{\textit{--- diverged ---}} \\
    \checkmark & -- & \checkmark & -- && 2.71 & 3.46 && 2.78 & 8.47\\
    \checkmark & \checkmark & \checkmark & -- && 2.15 & 2.99 && 2.45 & 7.53\\
    \checkmark & \checkmark & \checkmark & \checkmark && {\bf 1.99} & {\bf 2.80} && {\bf 2.01} & {\bf 6.72}\\
    \end{tabular}}
\vspace{0.0mm}
\caption{Ablation of proposed improvements. SQ: sequence loss, AU: heavy augmentation, FW: full-image warping, MF: multi-frame self-supervision. All components significantly improve performance and the sequence loss prevents divergence. 
}
\label{table:ablation_contributions}
\end{table}

\subsection{Ablation Study}
\label{subsec:ablation}

To determine which aspects of our model are responsible for its improved performance over prior work, we perform an extensive ablation study. In these ablations, we always train one model per domain (on KITTI-2015-test and Sintel-test after pretraining on Flying Chairs), and evaluate those on the corresponding validation split of the same domain.

\tightparagraph{RAFT Model}

Our first ablation investigates how much improvement can be obtained by taking the prior state of the art method UFlow and replacing its PWC model with RAFT. As the results in \autoref{table:comparison_pwc_raft} show, replacing the model without additional changes to the unsupervised learning method surprisingly does not improve but instead decreases performance. Through extensive experimentation, we identified and added the techniques presented here that enable superior unsupervised learning with RAFT. The gains from these techniques are much smaller with the PWC model, potentially because of the more constrained architecture.

\begin{table}[t]
\centering
\resizebox{\linewidth}{!}{
\begin{tabular}{lcccccc}
& \phantom{a}& \multicolumn{2}{c}{Sintel \textit{train}} & \phantom{a} & \multicolumn{2}{c}{KITTI-15 \textit{train}} \\
\cmidrule{3-4} \cmidrule{6-7} 
Self-sup. variant && Clean & Final && EPE & ER\%\\
\hline
No self-supervision && 3.88 & 4.50 && 3.22 & 8.40 \\
From intermediate predictions && 2.51 & 3.22 && 2.58 & 7.54 \\
W/ FB masking && 2.53 & 3.31 && 2.94 & 8.14 \\ %
W/o FB masking (Ours) && {\bf 2.15} & {\bf 2.99} && {\bf 2.45} & {\bf 7.53} \\
\end{tabular}}
\vspace{0.0mm}
\caption{Ablation of self-supervision improvements. All results are without multi-frame self-supervision but including all other components. FB masking refers to forward-backward consistency masking that prior work used in the self-supervision loss. Occlusions in the photometric loss are always masked.}
\label{table:ablation_selfsup}
\end{table}

\begin{figure*}[t]
    \centering
    \resizebox{\linewidth}{!}{
    \includegraphics{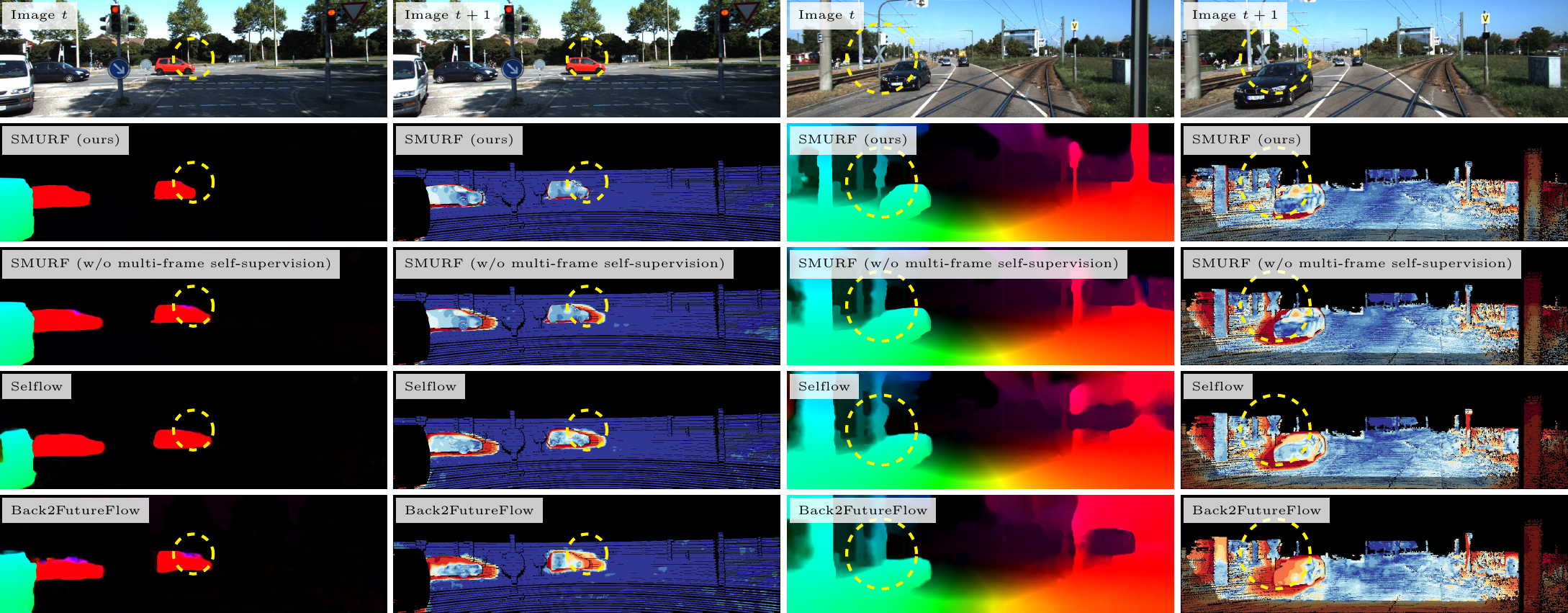}
    }
    \caption{Qualitative ablation and comparison of multi-frame self-supervision. The highlighted areas indicate clear improvements in occluded areas when using multi-frame self-supervision (second vs. third row). These improvements substantially outperform other unsupervised multi-frame methods (fourth and fifth row) although those use additional frames not only during training but also for inference.}
    \label{fig:kitti_benchmark_multiframe}
\end{figure*}

\newcommand{\resultswidththree}{2.22in}
\newcommand{\resultswidthtwo}{3.33in}
\newcommand{\resultswidthfour}{1.6in}

\newcommand{\resultsspacethree}{\,\,}
\newcommand{\resultsspacetwo}{\,}
\newcommand{\resultsspacefour}{\,}

\begin{figure}[t]
\centering
\begin{tabular}{@{}c@{\,\,}c@{\resultsspacefour}c@{\resultsspacefour}c@{}}
\includegraphics[width = \resultswidthfour]{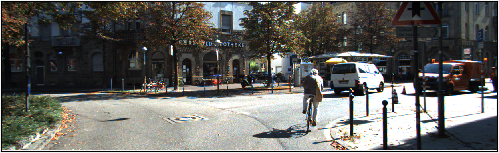} &
\includegraphics[width = \resultswidthfour]{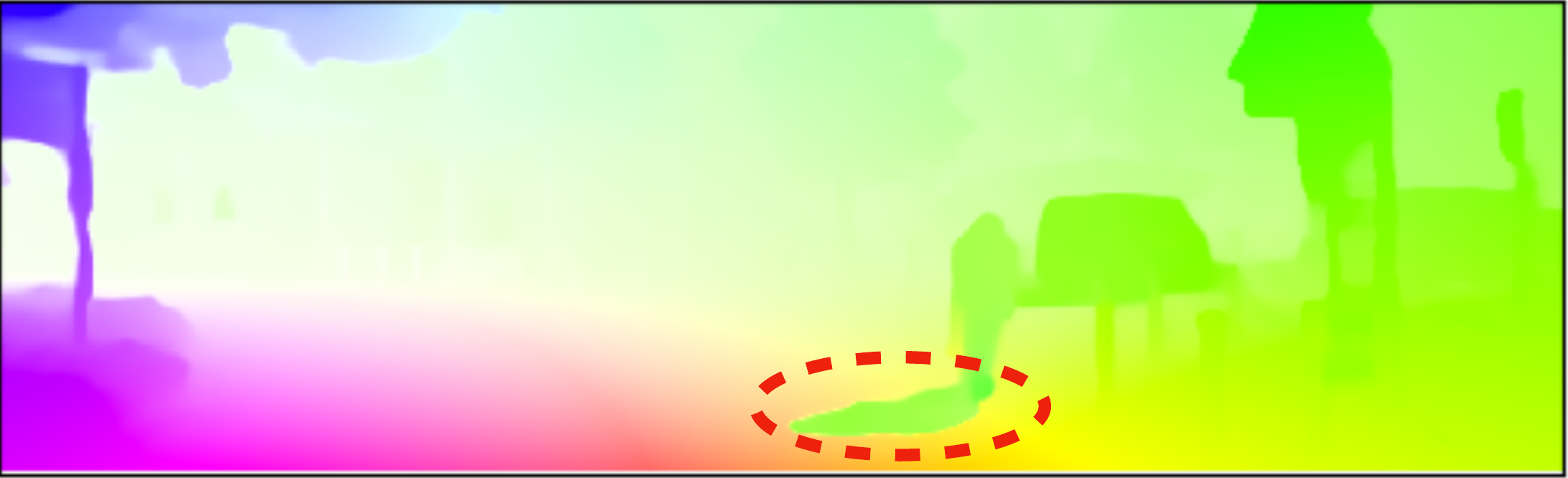} \\
 {\footnotesize Input RGB (first image)} & {\footnotesize Predicted Flow}
\end{tabular}
\caption{Limitations of unsupervised flow. Optimizing photometric consistency can produce incorrect flow at shadows / reflections.}
\label{fig:limitations}
\end{figure}

\tightparagraph{SMURF Components}

Next, we test different combinations of the novel components in our method. The results in \autoref{table:ablation_contributions} show that every component has a significant impact on performance. Sequence losses and heavy augmentations are necessary to achieve good results, especially on Sintel. And when using heavy augmentations, we need to apply sequence losses to prevent divergence of the model. Full-Image warping and multi-frame self-supervision have the strongest effect on KITTI, which makes sense as larger camera motion in that dataset causes more occlusions at the image boundaries which these components help to address. 

\tightparagraph{Self-Supervision Modifications}

We also ablated our proposed changes to self-supervision (\autoref{sec:unsupervised_raft}). The ablations in \autoref{table:ablation_selfsup} show that even with full-image warping, self-supervision remains an important component, and that it works best when not masking the loss as in prior work and when we use the final output (not intermediate model predictions) to generate the self-supervision labels.

\tightparagraph{Multi-Frame Self-Supervision}

Lastly, we provide a qualitative ablation of multi-frame self-supervision (\autoref{sec:multi_frame_self_supervision}) and a comparison to other unsupervised methods that use multi-frame information~\cite{Janai2018ECCV,SelFlow}. \autoref{fig:kitti_benchmark_multiframe} shows that multi-frame self-supervision substantially improves flow accuracy in occluded areas and does this much better than related multi-frame methods while being the only approach that requires multiple frames only during training and not for inference.

\subsection{Limitations}

A major limitation of unsupervised optical flow is that it estimates apparent visual motion rather than motion of physical objects (see \autoref{fig:limitations}). Overcoming this limitation requires some form of supervision, reasoning about the 3-D space as in scene flow~\cite{vedula1999three}, reasoning about semantics, or a combination of these. Future work could try to transfer the techniques from our method to such approaches.

\section{Conclusion}

We have presented SMURF, an effective method for unsupervised learning of optical flow that reduces the gap to supervised approaches and shows excellent generalization across datasets and even to ``zero-shot'' depth estimation. SMURF brings key improvements, most importantly (1) enabling the RAFT architecture to work in an unsupervised setting via modifications to the unsupervised losses and data augmentation, (2) full-image warping for learning to predict out of frame motion, and (3) multi-frame self-supervision for improved flow estimates in occluded regions. We believe that these contributions are a step towards making unsupervised optical flow truly practical, so that optical flow models trained on unlabeled videos can provide high quality pixel-matching in domains without labeled data.

\FloatBarrier

\pagebreak

{\small
\bibliographystyle{ieee_fullname}
\bibliography{references}

\begin{thebibliography}{10}\itemsep=-1pt

\bibitem{ReversingTheCycle}
Filippo Aleotti, Fabio Tosi, Li Zhang, Matteo Poggi, and Stefano Mattoccia.
\newblock Reversing the cycle: Self-supervised deep stereo through enhanced
  monocular distillation.
\newblock In {\em ECCV}, 2020.

\bibitem{Brox04}
Thomas Brox, Andr{\'e}s Bruhn, Nils Papenberg, and Joachim Weickert.
\newblock High accuracy optical flow estimation based on a theory for warping.
\newblock In {\em ECCV}, 2004.

\bibitem{ButlerECCV2012}
Daniel~J. Butler, Jonas Wulff, Garrett~B. Stanley, and Michael~J. Black.
\newblock A naturalistic open source movie for optical flow evaluation.
\newblock In {\em ECCV}, 2012.

\bibitem{chen2016full}
Qifeng Chen and Vladlen Koltun.
\newblock Full flow: Optical flow estimation by global optimization over
  regular grids.
\newblock In {\em CVPR}, 2016.

\bibitem{FlowNet}
Alexey Dosovitskiy, Philipp Fischer, Eddy Ilg, Philip H{\"a}usser, Caner
  Haz{\i}rba{\c{s}}, Vladimir Golkov, Patrick {van der Smagt}, Daniel Cremers,
  and Thomas Brox.
\newblock Flownet: Learning optical flow with convolutional networks.
\newblock In {\em ICCV}, 2015.

\bibitem{Gibson1950}
James~J. Gibson.
\newblock {\em {The Perception of the Visual World}}.
\newblock Houghton Mifflin, 1950.

\bibitem{Hirschmueller2008}
H. Hirschm\"uller.
\newblock Stereo processing by semi-global matching and mutual information.
\newblock {\em IEEE Transactions on Pattern Analysis and Machine Intelligence},
  30(2):328--341, February 2008.

\bibitem{Horn1981}
Berthold K.~P. Horn and Brian~G. Schunck.
\newblock Determining optical flow.
\newblock {\em AI}, 1981.

\bibitem{Flownet2}
Eddy Ilg, Nikolaus Mayer, Tonmoy Saikia, Margret Keuper, Alexey Dosovitskiy,
  and Thomas Brox.
\newblock Flownet 2.0: Evolution of optical flow estimation with deep networks.
\newblock In {\em CVPR}, 2017.

\bibitem{im2020unsupervised}
Woobin Im, Tae-Kyun Kim, and Sung-Eui Yoon.
\newblock Unsupervised learning of optical flow with deep feature similarity.
\newblock In {\em ECCV}, 2020.

\bibitem{Janai2018ECCV}
Joel Janai, Fatma G{\"u}ney, Anurag Ranjan, Michael~J. Black, and Andreas
  Geiger.
\newblock Unsupervised learning of multi-frame optical flow with occlusions.
\newblock In {\em ECCV}, 2018.

\bibitem{jonschkowski2020matters}
Rico Jonschkowski, Austin Stone, Jonathan~T Barron, Ariel Gordon, Kurt
  Konolige, and Anelia Angelova.
\newblock What matters in unsupervised optical flow.
\newblock In {\em ECCV}, 2020.

\bibitem{KingmaAdam}
Diederick~P Kingma and Jimmy Ba.
\newblock Adam: A method for stochastic optimization.
\newblock In {\em ICLR}, 2015.

\bibitem{liu2020learning}
Liang Liu, Jiangning Zhang, Ruifei He, Yong Liu, Yabiao Wang, Ying Tai, Donghao
  Luo, Chengjie Wang, Jilin Li, and Feiyue Huang.
\newblock Learning by analogy: Reliable supervision from transformations for
  unsupervised optical flow estimation.
\newblock In {\em CVPR}, 2020.

\bibitem{flow2stereo}
Pengpeng Liu, Irwin King, Michael Lyu, and Jia Xu.
\newblock {Flow2Stereo}: Effective self-supervised learning of optical flow and
  stereo matching.
\newblock In {\em CVPR}, 2020.

\bibitem{DDFlow}
Pengpeng Liu, Irwin King, Michael~R. Lyu, and Jia Xu.
\newblock {DDF}low: Learning optical flow with unlabeled data distillation.
\newblock In {\em AAAI}, 2019.

\bibitem{SelFlow}
Pengpeng Liu, Michael~R. Lyu, Irwin King, and Jia Xu.
\newblock Selflow: Self-supervised learning of optical flow.
\newblock {\em CVPR}, 2019.

\bibitem{lucas1981iterative}
Bruce~D. Lucas and Takeo Kanade.
\newblock An iterative image registration technique with an application to
  stereo vision.
\newblock {\em DARPA Image Understanding Workshop}, 1981.

\bibitem{maurer2018proflow}
Daniel Maurer and Andr{\'e}s Bruhn.
\newblock Proflow: Learning to predict optical flow.
\newblock In {\em BMVC}, 2018.

\bibitem{FlyingThings}
Nikolaus Mayer, Eddy Ilg, Philip Hausser, Philipp Fischer, Daniel Cremers,
  Alexey Dosovitskiy, and Thomas Brox.
\newblock A large dataset to train convolutional networks for disparity,
  optical flow, and scene flow estimation.
\newblock In {\em CVPR}, 2016.

\bibitem{meister2018unflow}
Simon Meister, Junhwa Hur, and Stefan Roth.
\newblock Unflow: Unsupervised learning of optical flow with a bidirectional
  census loss.
\newblock In {\em AAAI}, 2018.

\bibitem{KITTI2015}
Moritz Menze, Christian Heipke, and Andreas Geiger.
\newblock Joint 3d estimation of vehicles and scene flow.
\newblock {\em ISPRS Workshop on Image Sequence Analysis}, 2015.

\bibitem{spynet2017}
Anurag Ranjan and Michael~J. Black.
\newblock Optical flow estimation using a spatial pyramid network.
\newblock In {\em CVPR}, 2017.

\bibitem{ranjan2019cvpr}
Anurag Ranjan, Varun Jampani, Lukas Balles, Kihwan Kim, Deqing Sun, Jonas
  Wulff, and Michael~J. Black.
\newblock Competitive collaboration: Joint unsupervised learning of depth,
  camera motion, optical flow and motion segmentation.
\newblock In {\em CVPR}, 2019.

\bibitem{ren2017unsupervised}
Zhe Ren, Junchi Yan, Bingbing Ni, Bin Liu, Xiaokang Yang, and Hongyuan Zha.
\newblock Unsupervised deep learning for optical flow estimation.
\newblock In {\em AAAI}, 2017.

\bibitem{steinbrucker2009large}
Frank Steinbr{\"u}cker, Thomas Pock, and Daniel Cremers.
\newblock Large displacement optical flow computation without warping.
\newblock In {\em ICCV}, 2009.

\bibitem{Sun2010}
Deqing Sun, Stefan Roth, and Michael~J. Black.
\newblock Secrets of optical flow estimation and their principles.
\newblock In {\em CVPR}, 2010.

\bibitem{Sun2018PWCNet}
Deqing Sun, Xiaodong Yang, Ming-Yu Liu, and Jan Kautz.
\newblock {PWC-Net}: {CNNs} for optical flow using pyramid, warping, and cost
  volume.
\newblock In {\em CVPR}, 2018.

\bibitem{RAFT}
Zachary Teed and Jia Deng.
\newblock Raft: Recurrent all pairs field transforms for optical flow.
\newblock In {\em ECCV}, 2020.

\bibitem{tomasi1998bilateral}
Carlo Tomasi and Roberto Manduchi.
\newblock Bilateral filtering for gray and color images.
\newblock In {\em ICCV}, 1998.

\bibitem{instancenorm}
Dmitry Ulyanov, Andrea Vedaldi, and Victor~S. Lempitsky.
\newblock Instance normalization: The missing ingredient for fast stylization.
\newblock {\em arXiv:1607.08022}, 2016.

\bibitem{vedula1999three}
Sundar Vedula, Simon Baker, Peter Rander, Robert Collins, and Takeo Kanade.
\newblock Three-dimensional scene flow.
\newblock In {\em CVPR}, 1999.

\bibitem{wang2018unos}
Yang Wang, Peng Wang, Zhenheng Yang, Chenxu Luo, Yi Yang, and Wei Xu.
\newblock Unos: Unified unsupervised optical-flow and stereo-depth estimation
  by watching videos.
\newblock In {\em CVPR}, 2019.

\bibitem{UnOS}
Yang Wang, Peng Wang, Zhenheng Yang, Chenxu Luo, Yi Yang, and Wei Xu.
\newblock Unos: Unified unsupervised optical-flow and stereo-depth estimation
  by watching videos.
\newblock In {\em CVPR}, 2019.

\bibitem{wang2018occlusion}
Yang Wang, Yi Yang, Zhenheng Yang, Liang Zhao, Peng Wang, and Wei Xu.
\newblock Occlusion aware unsupervised learning of optical flow.
\newblock In {\em CVPR}, 2018.

\bibitem{xu2017accurate}
Jia Xu, Ren{\'e} Ranftl, and Vladlen Koltun.
\newblock Accurate optical flow via direct cost volume processing.
\newblock In {\em CVPR}, 2017.

\bibitem{yang2019volumetric}
Gengshan Yang and Deva Ramanan.
\newblock Volumetric correspondence networks for optical flow.
\newblock In {\em NeurIPS}, 2019.

\bibitem{yin2018geonet}
Zhichao Yin and Jianping Shi.
\newblock Geonet: Unsupervised learning of dense depth, optical flow and camera
  pose.
\newblock In {\em CVPR}, 2018.

\bibitem{jjyu2016unsupflow}
Jason~J. Yu, Adam~W. Harley, and Konstantinos~G. Derpanis.
\newblock Back to basics: Unsupervised learning of optical flow via brightness
  constancy and motion smoothness.
\newblock In {\em ECCV Workshop}, 2016.

\bibitem{zabih1994non}
Ramin Zabih and John Woodfill.
\newblock Non-parametric local transforms for computing visual correspondence.
\newblock In {\em ECCV}, 1994.

\bibitem{SsSMnet2017}
Yiran Zhong, Yuchao Dai, and Hongdong Li.
\newblock Self-supervised learning for stereo matching with self-improving
  ability.
\newblock {\em arXiv:1709.00930}, 2017.

\bibitem{Zhong2019UnsupervisedDE}
Yiran Zhong, Pan Ji, Jianyuan Wang, Yuchao Dai, and Hongdong Li.
\newblock Unsupervised deep epipolar flow for stationary or dynamic scenes.
\newblock In {\em CVPR}, 2019.

\bibitem{zou2018dfnet}
Yuliang Zou, Zelun Luo, and Jia-Bin Huang.
\newblock {DF}-{N}et: Unsupervised joint learning of depth and flow using
  cross-task consistency.
\newblock In {\em ECCV}, 2018.

\end{thebibliography}
}

\end{document}